%% file: main.tex
\documentclass[10pt,twocolumn,letterpaper]{article}

\usepackage{iccv}
\usepackage{times}
\usepackage{epsfig}
\usepackage{graphicx}
\usepackage{amsmath}
\usepackage{amssymb}
\usepackage{subfig}
\usepackage{diagbox}
\usepackage{authblk}
\usepackage{blindtext}

\makeatletter
\renewcommand\AB@affilsepx{, \protect\Affilfont}
\makeatother

\usepackage[pagebackref=true,breaklinks=true,letterpaper=true,colorlinks,bookmarks=false]{hyperref}

\iccvfinalcopy % *** Uncomment this line for the final submission

\def\iccvPaperID{3273} % *** Enter the ICCV Paper ID here

\ificcvfinal\fi
\begin{document}

%%%%%%%%% TITLE
\title{An Internal Learning Approach to Video Inpainting}

\author{Haotian Zhang$^1$
        \thanks{This work was done primarily during Haotian Zhang's internship at Adobe Research.}
        \quad Long Mai$^2$
        \quad Ning Xu$^2$ 
        \quad Zhaowen Wang$^2$
        \quad John Collomosse$^{2,3}$
        \quad Hailin Jin$^2$\\
        $^1$Stanford University
        \quad $^2$Adobe Research
        \quad $^3$University of Surrey\\
        \tt\small {haotianz@stanford.edu \quad\{malong,nxu,zhawang,collomos,hljin\}@adobe.com}
       }

\maketitle
%\thispagestyle{empty}

%%%%%%%%% ABSTRACT
\begin{abstract}

%jpc compacted
We propose a novel video inpainting algorithm that simultaneously hallucinates missing appearance and motion (optical flow) information, building upon the recent `Deep Image Prior' (DIP) that exploits convolutional network architectures to enforce plausible texture in static images.  In extending DIP to video we make two important contributions.  First, we show that coherent video inpainting is possible without a priori training.  We take a generative approach to inpainting based on internal (within-video) learning  without reliance upon an external corpus of visual data to train a one-size-fits-all model for the large space of general videos. Second, we show that such a framework can jointly generate both appearance and flow, whilst exploiting these complementary modalities to ensure mutual consistency. 
We show that leveraging appearance statistics specific to each video achieves visually plausible results whilst handling the challenging problem of long-term consistency.
% We show that leveraging appearance statistics specific to a video can achieve state-of-the-art results whilst handling the challenging problem of long-term consistency.

%We propose a novel video inpainting algorithm that simultaneously hallucinates missing appearance and motion (optical flow) information, building upon the recent  `deep image prior’ (DIP) that exploits convolutional neural network (CNN) architecture to enforce plausible texture in static images.  In extending DIP to video we make two important contributions.  First, we show that coherent video inpainting is possible with no a priori training.  We take a generative approach to inpainting based on internal (within-video) learning of CNN parameters without reliance upon an external corpus of visual data to train a one-size-fits-all model for the large space of general video. Second, we show that such a framework can be leveraged to jointly generate both appearance and flow, whilst exploiting these complementary modalities to ensure mutual consistency. Experiments on a wide range of videos collected from the DAVIS segmentation dataset and a composed dataset \textcolor{red}{HZ: Classic videos} demonstrate that our proposed algorithm can leverage appearance statistics specific to the video to achieve visually plausible results whilst handling the challenging problem of long-range consistency.
\end{abstract}

%%%%%%%%% BODY TEXT
\vspace{-5mm}
\section{Introduction}
\label{sec:intro}
\input{latex/Intro.tex}

\section{Related Work}
\label{sec:related_work}
\input{latex/Related_Work.tex}

\section{Video Inpainting via Internal Learning}
\label{sec:method}
\input{latex/Method.tex}

\section{Experiments}
\label{sec:experiments}
\input{latex/Experiments.tex}

\section{Discussion}
\label{sec:discussion}
\input{latex/Discussion.tex}

\section{Conclusion}
\label{sec:conclusion}

\input{latex/Conclusion.tex}

{\small

\input{latex/main.bbl}
}

\newpage
\appendix
\input{latex/Appendix.tex}

\end{document}

%% file: latex/Intro.tex
\begin{figure*}[t!]
\begin{center}
% \fbox{\rule{0pt}{2in} \rule{.9\linewidth}{0pt}}
\includegraphics[width=1\linewidth]{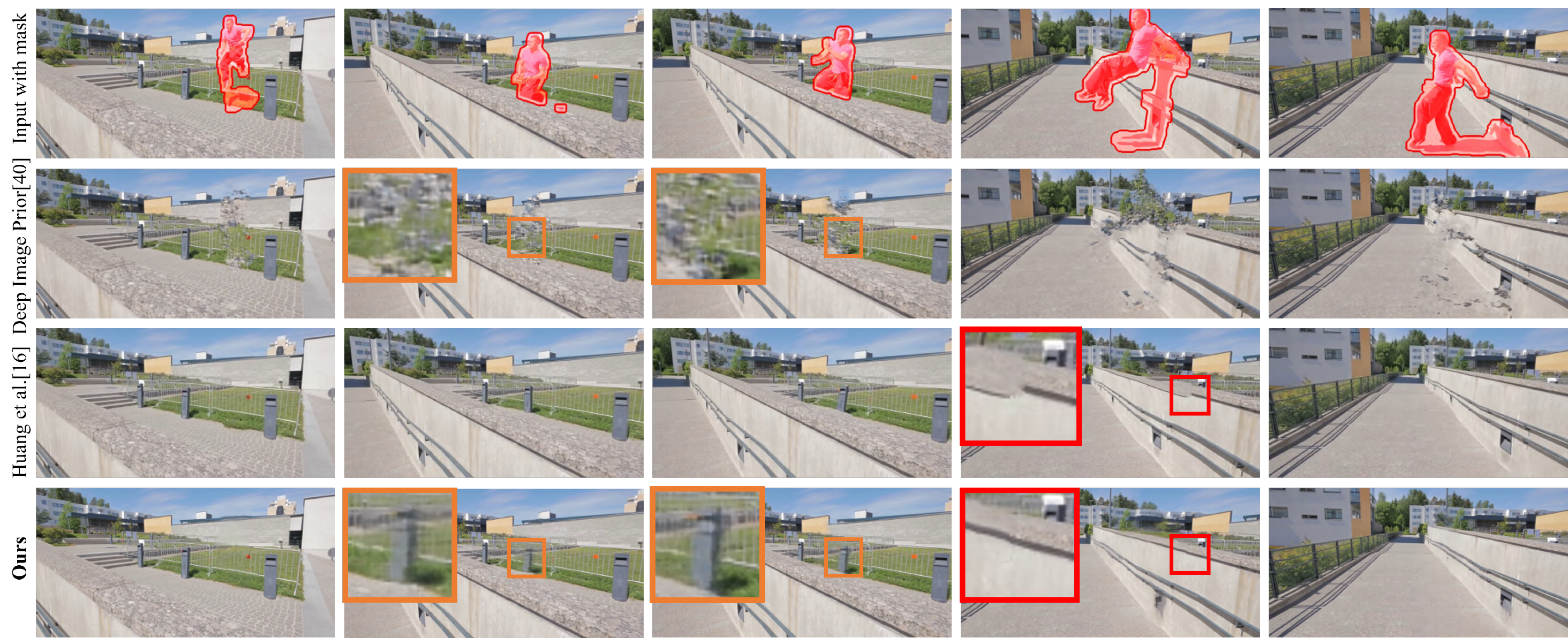}
\end{center}
\vspace{-4mm}
   \caption{Video inpainting results for the `parkour' sequence. Our video-based internal learning framework enables much more coherent video inpainting results compared to the frame-based baseline~\cite{Ulyanov_CVPR_2018} ($2^{nd}$ row), even for content unseen in multiple frames (orange box). As a network-based synthesis framework, our method can employ natural image priors to avoid shape distortions, which often occur in patch-based methods such as ~\cite{Huang_SA_2016} ($3^{rd}$ row) for challenging videos (red box).}
\label{fig:head}
\vspace{-5mm}
\end{figure*}

%Video inpainting is the problem of synthesizing plausible visual content within a missing region (`hole'); for example, to remove unwanted objects.  Video inpainting is fundamentally ill-posed; there is no unique solution for the missing content. Rather, the goal is to generate visually plausible content that is coherent in both space and time. Priors play a critical role in expressing these constraints. Traditional patch-based methods ~\cite{Wexler_CVPR_2004,Newson_SIAM_2014,Huang_SA_2016,Le_ICIP_2017} learn priors from hand-crafted metrics defined on the pixel domain, which is prone to shape distortions. Recent image inpainting approaches \cite{pathak_CVPR_2016,Iizuka_SIGGRAPH_2017,Yu_CVPR_2018, Liu_ECCV_2018} learn  priors from an external image corpus via a deep neural network, applying the learned appearance model to hallucinate content conditioned upon known regions. Extending these deep generative approaches to video is challenging for two reasons. First, the coherency constraints for video are much stricter than for images. The hallucinated content must be not only consistent within its own frame, but also consistent with earlier or later frames. Second, the space of video is orders of magnitude larger than that of images. It is difficult to train a single model on an external dataset to learn effective priors for general videos, as one requires not only a sufficiently expressive model to generate that space of variation, but also exponential volumes of data to provide sufficient  coverage.

Video inpainting is the problem of synthesizing plausible visual content within a missing region (`hole'); for example, to remove unwanted objects. Inpainting is fundamentally ill-posed; there is no unique solution for the missing content. Rather, the goal is to generate visually plausible content that is coherent in both space and time. Priors play a critical role in expressing these constraints. Patch-based optimization methods ~\cite{Huang_SA_2016,Le_ICIP_2017,Newson_SIAM_2014,Wexler_CVPR_2004} effectively leverage different priors such as patch recurrence, total variation, and motion smoothness to achieve state-of-the-art video inpainting results. These priors, however, are mostly hand-crafted and often not sufficient to capture natural image priors, which often leads to distortion in the inpainting results, especially for challenging videos with complex motion (Fig. ~\ref{fig:head}). Recent image inpainting approaches ~\cite{Iizuka_SIGGRAPH_2017,Liu_ECCV_2018,pathak_CVPR_2016,Yu_CVPR_2018} learn better image priors from an external image corpus via a deep neural network, applying the learned appearance model to hallucinate content conditioned upon observed regions. Extending these deep generative approaches to video is challenging for two reasons. First, the coherency constraints for video are much stricter than for images. The hallucinated content must not only be  consistent within its own frame, but also be consistent across adjacent frames. Second, the space of videos is orders of magnitude larger than that of images, making it challenging to train a single model on an external dataset to learn effective priors for general videos, as one requires not only a sufficiently expressive model to generate all variations in the space, but also large volumes of data to provide sufficient coverage.

This paper proposes {\em internal learning} for video inpainting inspired by the recently proposed `Deep Image Prior' (DIP) for single image generation ~\cite{Ulyanov_CVPR_2018}. The striking result of DIP is that  `knowledge' of natural images can be encoded through a convolutional neural network (CNN) architecture; \ie the network structure rather than actual filter weights. The translation equivariance of CNN enables DIP to exploit the internal recurrence of visual patterns in images ~\cite{Irani_CVPR_2018}, in a similar way as the classical patch-based approaches ~\cite{Glasner_2009_ICCV} but with more expressiveness. Furthermore, DIP does not require an external dataset and therefore suffers less from the aforementioned exponential data problem. We explore this novel paradigm of DIP for video inpainting as an alternative to learning priors from external datasets.

Our core technical contribution is the first internal learning framework for video inpainting. Our study establishes the significant result that it is possible to \textit{internally train a single frame-wise generative CNN to produce high quality video inpainting results}. We report on the effectiveness of different strategies for internal learning to address the fundamental challenge of temporal consistency in video inpainting. Therein, we develop a consistency-aware training strategy based on joint image and flow prediction. Our method enables the network to not only capture short-term motion consistency but also propagate the information across distant frames to effectively handle long-term consistency. We show that our method, whilst trained internally on one (masked) input video without any external data, can achieve state-of-the-art video inpainting result. As a network-based framework, our method can incorporate natural image priors learned from CNN to avoid shape distortions which occur in patch-based methods. (Fig. ~\ref{fig:head})
% generate plausible and coherent video inpainting results for diverse and challenging footage.

A key challenge in extending DIP to video is to ensure temporal consistency;  content should be free from visual artifacts and exhibit smooth motion (optical flow) between adjacent frames.  This is especially challenging for video inpainting (\eg versus video denoising) due to the reflexive requirements of pixel correspondence over time to generate missing content, as well as such correspondence to enforce temporal smoothness of that content. We break this cycle by {\em jointly synthesizing content in both appearance and motion domains}, generating content through an Encoder-Decoder network that exploits DIP not only in the visual domain but also in the motion domain.  This enables us to jointly solve the inpainted appearance and optical flow field -- maintaining consistency between the two. We show that simultaneous prediction of both appearance and motion information not only enhances spatial-temporal consistency, but also improves visual plausibility by better propagating structural information within larger hole regions.

%% file: latex/Related_Work.tex
\noindent\textbf{Image/Video Inpainting}.
The problem of image inpainting/completion~\cite{survey_paper} has been studied extensively, with classical approaches focusing on patch-based non-parametric optimization~\cite{Barnes_SIGGRAPH_2009, Hays_SIGGRAPH_2007, He_PAMI_2014, Lee_CVPR_2016, Huang_SIGGRAPH_2014, Kalantari_ICCP_2014, Komodakis_CVPR_2006, Le_Meur_TIP_2013, Sun_SIGGRAPH_2005, Xu_TIP_2010} as well as more recent work using deep generative neural networks~\cite{Iizuka_SIGGRAPH_2017, Liu_ECCV_2018, pathak_CVPR_2016, Yang_CVPR_2017, Yu_CVPR_2018}. On the other hand, the video inpainting problem has received far less attention from research community.
Most existing video inpainting methods build on patch-based synthesis with spatial-temporal matching~\cite{Huang_SA_2016,Le_ICIP_2017,Newson_SIAM_2014,Wexler_CVPR_2004} or explicit motion estimation and tracking~\cite{Afifi_ISPACS_2014,ebdelli2015video,granados2012background,granados2012not}. Very recently, deep convolutional networks have been used to directly inpaint holes in videos and achieve promising results~\cite{kim2019deep,wang2018video,xu2019deepflow}, leveraging large external video corpus for training along with specialized recurrent frameworks to model spatial-temporal coherence. Different from their works, we explore the orthogonal direction of learning-based video inpainting by investigating an internal (within-video) learning approach. Video inpainting has also been used as a self-supervised task for deep feature learning~\cite{nallabolu2017unsupervised} which has a different goal from ours.
% In this work, we expand the horizon of deep network based video inpainting by investigating an internal (within-video) learning approach without training on any external visual corpus.
%We show for the first time that a single 2D-CNN generative model can produce highly plausible video inpainting results from pure noise as input for each frame, while synthesizing each frame independently. 

\noindent\textbf{Internal Learning}.
% Our work is inspired by the recent Deep Image Prior (DIP) work by Ulyanov \etal~\cite{Ulyanov_CVPR_2018} which shows that a static image may be inpainted by a CNN-based generative model trained directly on the non-hole region of the same image with a reconstruction loss. The trained model encodes the visible image contents with white noise, which at the same time enables the synthesis of plausible texture in the hole region.
% Our work is the first to extend DIP to video, providing an in-depth exploration of effective internal learning strategies for video inpainting. Our study demonstrates that with a properly designed internal learning algorithm, it is possible to inpaint hundreds of video frames with a single shared 2D CNN network. 
Our work is inspired by the recent `Deep Image Prior' (DIP) work by Ulyanov \etal~\cite{Ulyanov_CVPR_2018} which shows that a static image may be inpainted by a CNN-based generative model trained directly on the non-hole region of the same image with a reconstruction loss. The trained model encodes the visible image contents with white noise, which at the same time enables the synthesis of plausible texture in the hole region. The idea of internal learning has also been shown effective in other application domains, such as image super-resolution ~\cite{Irani_CVPR_2018}, semantic photo manipulation ~\cite{Bau:Ganpaint:2019} and video motion transfer ~\cite{chan2018everybody}.
Recently, Gandelsman \etal~\cite{gandelsman2018double} further proposes `Double-DIP' for unsupervised image decomposition by reconstructing different layers with multiple DIP. Their framework can also be applied for video segmentation. In this paper, we extend a single DIP to video and explore the effective internal learning strategies for video inpainting.
% Our study demonstrates that with a properly designed internal learning algorithm, it is possible to inpaint hundreds of video frames with a single shared 2D CNN network. 

\noindent\textbf{Flow Guided Image/Video Synthesis}.
Only encoding frames with a 2D CNN is insufficient to maintain the temporal consistency of a video. Conventionally, people have used optical flow from input source videos as guidance to enhance the temporal consistency of target videos in various video processing tasks such as denoising~\cite{Ji2010}, super-resolution~\cite{huang2018superres}, frame interpolation~\cite{nvidiacvpr2018}, and style transfer~\cite{gupta2017characterizing}.
Our work incorporates the temporal consistency constraint of inpainted area by jointly generating images and flows with a new loss function.

%% file: latex/Method.tex
\begin{figure}[t!]
\begin{center}
% \fbox{\rule{0pt}{2in} \rule{.9\linewidth}{0pt}}
\includegraphics[width=0.99\linewidth]{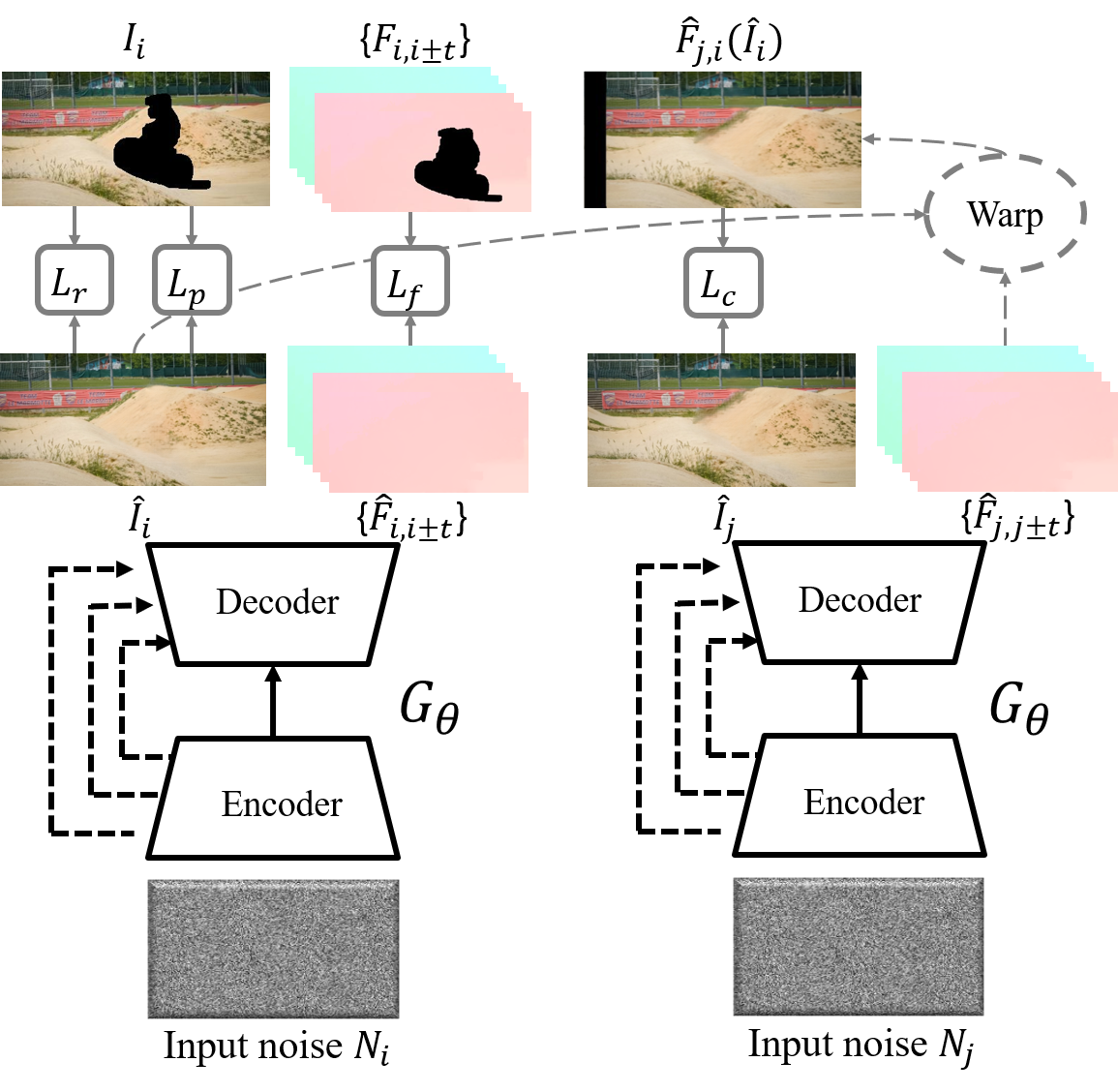}
\vspace{-5mm}
   \caption{Overview of our video inpainting framework. Given the input random noise $N_i$ for each individual frame, a generative network $G_{\theta}$ is used to predict both frame $\hat{I}_i$ and optical flow maps $\hat{F}_{i,i\pm t}$. $G_{\theta}$ is trained entirely on the input video (with holes) without any external data, optimizing the combination of the image generation loss $L_r$, perceptual loss $L_p$, flow generation loss $L_f$ and consistency loss $L_c$. }
\label{fig:framework}
\vspace{-8mm}
\end{center}
\end{figure}

The input to video inpainting  is a (masked) video sequence $\bar{V} = \{I_i \odot M_i\}_{i=1..T}$ where $T$ is the total number of frames in the video. $M_i$ is the binary mask defining the known regions in each frame $I_i$ (1 for the known regions, and 0 otherwise). $\odot$ denotes the element-wise product. Let $I^*_i$ denote the desired version of $I_i$ where the masked region is filled with the appropriate content. The goal in video inpainting is to recover  $V^* = \{I^*_i\}_{i=1..T}$ from $\bar{V}$. 

In this work, we approach video inpainting with an internal learning formulation. The general idea is to use $\bar{V}$ as the training data to learn a generative neural network $G_{\theta}$ to generate each target frame $I^*_i$ from a corresponding noise map $N_i$. The noise map $N_i$ has one channel and shares the same spatial size with the input frame. We sample the input noise maps independently for each frame and fix them during training. Once trained, $G_{\theta}$ can be used to generate all the frames in the video to produce the inpainting results. 
\begin{small}
\begin{equation}
\vspace{-1mm}
    I^*_i = G_{\theta^*}(N_i)
\vspace{-2mm}
\end{equation}
\end{small}
\!\!where $\theta$ denotes the network parameters which are optimized during the training process. We implement $G_{\theta}$ as an Encoder-Decoder architecture with skip connections. For each input video, we train an individual model from scratch. 

One may concern that a generative model $G_{\theta}$ defined in this way would be too limited for the task of video inpainting as it does not contain any temporal modeling structure required for video generation, \eg recurrent prediction, attention, memory modeling, \etc. In this paper, however, we intentionally keep this extreme form of internal learning and focus on exploring appropriate learning strategies to unleash its potential to perform the video inpainting task. In this section, we discuss our training strategies to train $G_{\theta}$ such that it can generate plausible $V^*$.

\subsection{Loss Functions}
\label{sec:loss_function}

Let $\hat{I}_i = G_{\theta}(N_i)$ be the network output at frame $i$. We define a loss function $L$ at each frame prediction $\hat{I}_i$ and accumulate the loss over the whole video to obtain the total loss  to optimize the network parameters during training.
\begin{small}
\vspace{-2mm}
\begin{equation}
\label{eq:total_loss_function}
    L = \omega_{r} L_r + \omega_{f} L_f + \omega_c L_c + \omega_p L_p 
\end{equation}
\vspace{-1mm}
\end{small}
\kern-0.75em where $L_r$, $L_f$, $L_c$, and $L_p$ denote the image generation loss, flow generation loss, consistency loss, and perceptual loss, respectively. The weights are empirically set as $\omega_{r}{=}1$, $\omega_{f}{=}0.1$, $\omega_{c}{=}1$, $\omega_{p}{=}0.01$ and fixed in all of our experiments. We define each individual loss term as follows: 

\noindent\textbf{Image Generation Loss}.
In the context of image inpainting, ~\cite{Ulyanov_CVPR_2018} employs the $L_2$ reconstruction loss defined on the known regions of the image. Our first attempt to explore internal learning for video inpainting is to define a similar  generation loss on each predicted frame.
\begin{small}
\vspace{-1mm}
\begin{equation}
\label{eq:image_loss}
    L_{r} (\hat{I_i}) = \parallel M_i\odot \big( \hat{I_i} -  I_i \big) \parallel^2_2
\end{equation}
\end{small}
\noindent\textbf{Flow Generation Loss}.
Image generation loss enables the network to reconstruct individual frames, but fails to capture the temporal consistency across frames. Therefore, it is necessary to allow information to be propagated across frames. Our key idea is to encourage the network to learn such propagation mechanism during training. 
%To this end, we incorporate an intuitive yet critical extension to our internal learning framework. 
We first augment the network to jointly predict the color and flow values at each pixel: $(\hat{I}_i, \hat{F}_{i,j}) = G_{\theta}(N_i)$, where $\hat{F}_{i,j}$ denotes the predicted optical flow from frame $i$ to frame $j$ (Fig.~\ref{fig:framework}). To increase the robustness and better capture long-term temporal consistency, our network is designed to jointly predict flow maps with respect to 6 adjacent frames of varying temporal directions and ranges: $j \in \{i \pm 1, i \pm 3, i \pm 5\}$. We define the flow generation loss similarly as the image generation loss to encourage the network to learn the `flow priors' from the known regions:
\begin{small}
\begin{equation}
\label{eq:flow_loss}
    L_f (\hat{F}_{i,j}) = \parallel O_{i,j} \odot M^f_{i,j} \odot \big( \hat{F}_{i,j} -  F_{i,j} \big) \parallel^2_2.
\end{equation}
\end{small}
\noindent The known flow $F_{i,j}$ is estimated using PWC-NET~\cite{sun2018pwc} from the original input frame $I_i$ to  $I_j$, which also estimates the occlusion map $O_{i,j}$ through the forward-backward consistency check. $M^f_{i,j} = M_i \cap M_j(F_{i,j})$ represents the reliable flow region computed as the intersection of the aligned masks of frame $i$ and $j$.

%With the network now jointly predicts visual frames and flows, we define the simple flow-image consistency loss to encourage the network to capture the useful interaction between images and flows: the generated flow maps and the generated neighboring frames need to be consistent. \textcolor{red}{not very clear, the generated flow maps and the generated frames have constraints on each other}. 

\begin{figure}[t!]
\begin{center}
\includegraphics[width=1.0\linewidth]{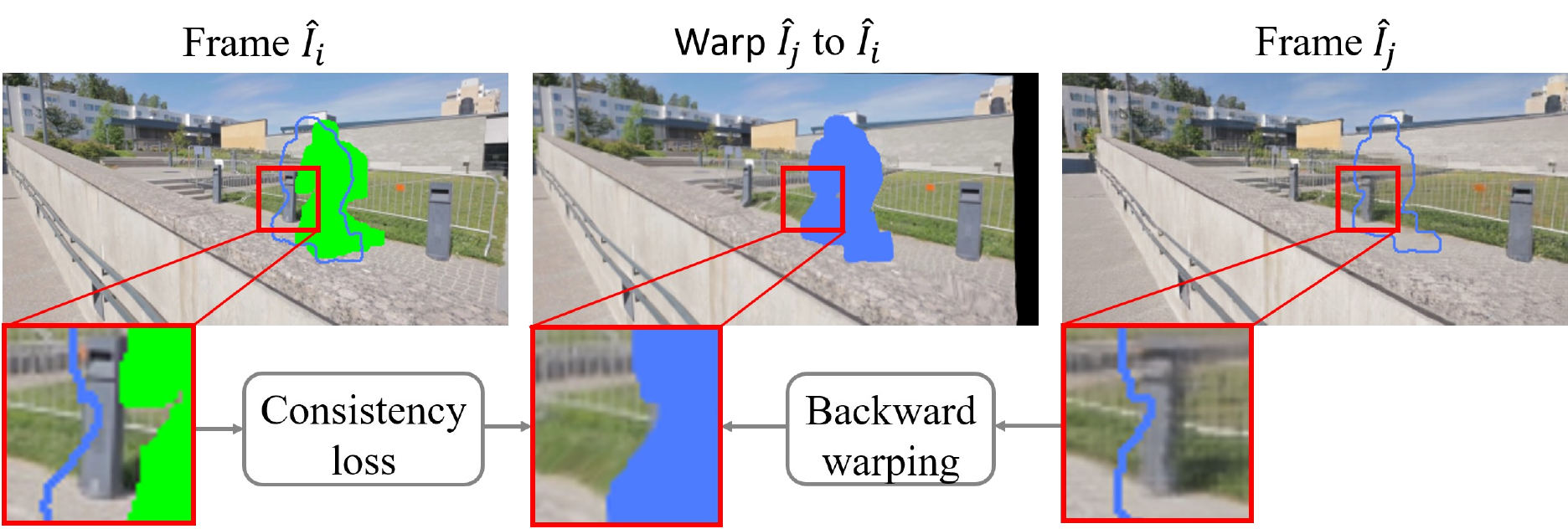}
\end{center}
\vspace{-4mm}
   \caption{Effectiveness of consistency loss. By warping the predicted frame $\hat{I_j}$ into frame $\hat{I_i}$, part of the  hole region in $\hat{I_j}$ can be spatially matched to the visible regions in $\hat{I_i}$ (red boxes). This not only provides useful training signal to constrain the inpainting of that region in $\hat{I_j}$, but also effectively propagates the content in the visible regions from one frame into the hole regions of its neighboring frames.}
\label{fig:consistency_loss}
\vspace{-5mm}
\end{figure}

\noindent \textbf{Consistency Loss}.
With the network jointly predicts images and flows, we define the image-flow consistency loss to encourage the generated frames and the generated flows to constrain each other: the neighboring frames should be generated such that they are consistent with the predicted flow between them.
\begin{small}
\vspace{-2mm}
\begin{equation}
\label{eq:consistency_loss}
    L_{c} (\hat{I}_{j}, \hat{F}_{i,j}) =  \parallel 
    (1-M^f_{i,j}) \odot \big( \hat{I}_{j}(\hat{F}_{i,j}) -
    \hat{I}_{i} \big) \parallel^2_2
\end{equation}
\end{small}
\noindent where $\hat{I}_{j}(\hat{F}_{i,j})$ denotes the warped version of the generated frame $\hat{I}_j$ using the generated flow $\hat{F}_{i,j}$ through backward warping. We constrain this loss only in the hole regions using the inverse mask $1{-}M^f_{i,j}$ to encourage the training to focus on propagating information inside the hole. We find this simple and intuitive loss term allows the network to learn the notion of flow and leverage it to propagate training signal across distant frames (as illustrated in Fig.~\ref{fig:consistency_loss}).

%We temporarily fix $\hat{I}_i$ here and only update on $\hat{I}_j$ and $\hat{F}_{i,j}$. 

%This loss encourages the two predicted hole regions from frames $I_i$ and $I_j$ to be consistent according to the generated flow $\hat{F}_{i,j}$ between them. 

%Like the image prediction, we will also apply L2 loss for flow on the non-hole region, besides non-occlusion region. 

%We define the simple flow-image consistency loss to encourage the network to capture the useful interaction of flows and images: the generated flow maps and the generated neighboring frames need to be consistent. To define this loss term, we use the predicted flow $F_{i,j}$ to warp the predicted frame $I_{j}$ back to time t, and then measures the L2 distance between the warped frame and the predicted frame $I_i$ only inside the hole region. 

% In our experiments, we make use of three flow predictions in both forward and backward directions at each frame, corresponding to $j \in \{i \pm 1, i \pm 3, i \pm 5\}$. 
%To leverage long-range temporal consistency, our model generates flows at different frame intervals in both forward and backward directions.
%i.e., given frame $i$, we calculate flows between frames $j \in \{i \pm 1, i \pm 3, i \pm 5\}$.
%We found this simple and intuitive loss term allows the network to learn useful notion of flows and is capable of using that to propagate training signal across distant frames (as illustrated in Fig.~\ref{fig:consistency_loss}) 

%Note that this loss term only encodes local consistency information, no explicit flow tracking or temporal-aware patch matching is encoded in the network. 

%\textcolor{red}{Todo: add another graph showing the propagation process by measuring PSNR or SSIM for a fixed batch }

%We found that this loss is critical to constrain the unknown region in frame $I_{j}$ and flow $F_{i,j}$.  

%\textcolor{red}{Fig.}~\ref{fig:consistency_loss_effect} shows the effect of adding incorporating the consistency loss. Without this loss, the model can make use of spatial context to synthesize plausible results frame-wise. However, the information is inconsistent with the same information that already appeared in the earlier and later frames. The network trained with our consistency loss can capture how local content moves across frame and learn to reconstruct the right re-appearing content in distant frame.

\noindent\textbf{Perceptual Loss}.
%One important advantage of CNN-based reparameterization is that it makes the entire optimization problem differentiable. That makes it easy to incorporate useful priors from any external network pre-trained on large-scale datasets. As an example, we incorporate the popular perceptual loss, defined according to the feature similarity on layers from the pre-trained VGG model~\cite{johnson2016perceptual}.
To further improve the frame generation quality, we incorporate the popular perceptual loss, defined according to the similarity on extracted feature maps from the pre-trained VGG16 model~\cite{johnson2016perceptual}.

%This loss function encourages the generated frame $F_{\theta}(N_i)$ and the ground-truth $V_i$ to have similar feature representations as computed by a pretrained feature extraction network $\phi$. $\phi_j(x)$ denotes the feature map of the $j$th layer given input $x$. 
\begin{small}
\vspace{-3mm}
\begin{equation}
    L_p(\hat{I_i}) = \Sigma_{k \in K} \parallel \psi_k(M_i) \odot \big( \phi_k(\hat{I_i}) -  \phi_k(I_i) \big) \parallel^2_2
\end{equation}
\end{small}
\noindent where $\phi_k(I_i)$ denotes the feature extracted from  $I_i$ using the $k^{th}$ layer of the pre-trained VGG16 network, $\psi_k(M_i)$ denotes the resized mask with the same spatial size as the feature map.  This perceptual loss has been used to improve the visual sharpness of  generated images~\cite{johnson2016perceptual,Ledig_CVPR_2017,Niklaus_CVPR_2017,Zhang_2018_CVPR}. We use 3 layers $ \{\mbox{relu}1\_2, \mbox{relu}2\_2, \mbox{relu}3\_3\}$ to define our perceptual loss.

\subsection{Network Training}

% As in standard neural network training procedures, we train our network in a stochastic manner. In each iteration through a video, we randomly select a batch of five frames and update the network parameters using back-propagation  with the gradients obtained from local loss value defined in Equation \ref{eq:total_loss_function} computed from the batch. In our experiments, we train our network for 20 epochs (where one epoch corresponds to a full pass through the video).

While the standard stochastic training works reasonably, we use the following curriculum-based training procedure during network optimization: Instead of using pure random frames in one batch, we pick $N$ frames which are consecutive with a fixed frame interval of $t$ as a batch. While training with the batch, the flow generation loss and consistency loss are computed only using the corresponding flows ($F_{i,i\pm t}$). We find this helps propagate the information more consistently across the frames in the batch. In addition, inspired by DIP, we perform the parameter update for each batch multiple times continuously, with one forward pass and one back-propagation each time. This allows the network to be optimized locally until the image and flow generation reach their consistent state. We find that using 50-100 updates per batch gives the best performance through experiments of hyper-parameter tuning.
% While the standard stochastic training works reasonably, we find the following curriculum-based training procedure more effective in encouraging information propagation during network optimization: Instead of using pure random frames in one batch, we randomly pick $n$ frames which are consecutive with a fixed frame interval of $t$ as a batch to feed in the network and compute the flow-related loss (Eq.~\ref{eq:flow_loss} and Eq.~\ref{eq:consistency_loss}) using only the corresponding flows ($F_{i,i\pm t}$). We find this helps propagate the information more consistently across the frames in the batch. In addition, instead of performing parameter update once for each batch, we perform the update multiple times. This allows the network to be optimized locally until the image and flow generation reach their consistent state. We find that using 50-100 updates per batch gives the best performance.

%In our experiments, we train our network for 20 epochs (where one epoch corresponds to a full pass through the video). Within each epoch, we randomly pick five frames which are consecutive with the stride of 1, 3 or 5 as a batch to feed in the network, so that we can collect predicted flow with the same stride to compute consistency loss. 

\subsection{Implementation Details}

%One advantage of this framework is that the optimization process becomes conceptually simple: standard network training algorithms exists with robust performance and requires no domain-specific optimization tricks to work as in the case of existing works. 
%\textcolor{red}{Network structure, multi-flow, multi-batch}

We implement our method using  PyTorch  %package\footnote{https://pytorch.org/} 
and run our experiments on a single NVIDIA TITAN Xp GPU. We initialize the model weights using the initialization method described in ~\cite{lecun2012efficient} and use Adam ~\cite{kingma2014adam} with the learning rate of 0.01 and batch size of 5 during training. 

Our network is implemented as an Encoder-Decoder architecture with skip connections, which is found to perform well for image inpainting ~\cite{Ulyanov_CVPR_2018}. The details of the network architecture is provided in the supplementary material.
% Specifically, the network consists of 6 convolution blocks in both the encoder and the decoder, where each  block consists of 2 convolution layers, followed by a Batch-Norm layer and a LeakyReLU activation. The image and flow generation branches share all the convolution blocks except the final $1 \times 1$ convolution layer.
% The output of the flow prediction branch consists of 12 channels, corresponding to 6 flow predictions of temporal range 1, 3, 5 in both forward and backward directions.

%We also experiment with different architectures as in~\cite{Ulyanov_CVPR_2018} and do not find significant differences in performance. 
%for reaching their specific number of output channels. 

% In our experiment, we train our network for 1000 batches for each video. In each batch, we randomly pick five frames with a fixed stride of 1, 3 or 5, so that we can collect predicted flow with the same stride to compute consistency loss. We initialize the weights using the initialization method described in ~\cite{lecun2012efficient} and use Adam \cite{kingma2014adam} for optimization. We use the learning rate 0.01 with the weight $\omega_{ri} = 1$, $\omega_{rf} = 0.1$, $\omega_{c} = 1$, $\omega_{p} = 0.01$ across all examples.

%% file: latex/Experiments.tex
\begin{figure*}[ht]\vspace{-0.2in}
\begin{center}
\includegraphics[width=1\linewidth]{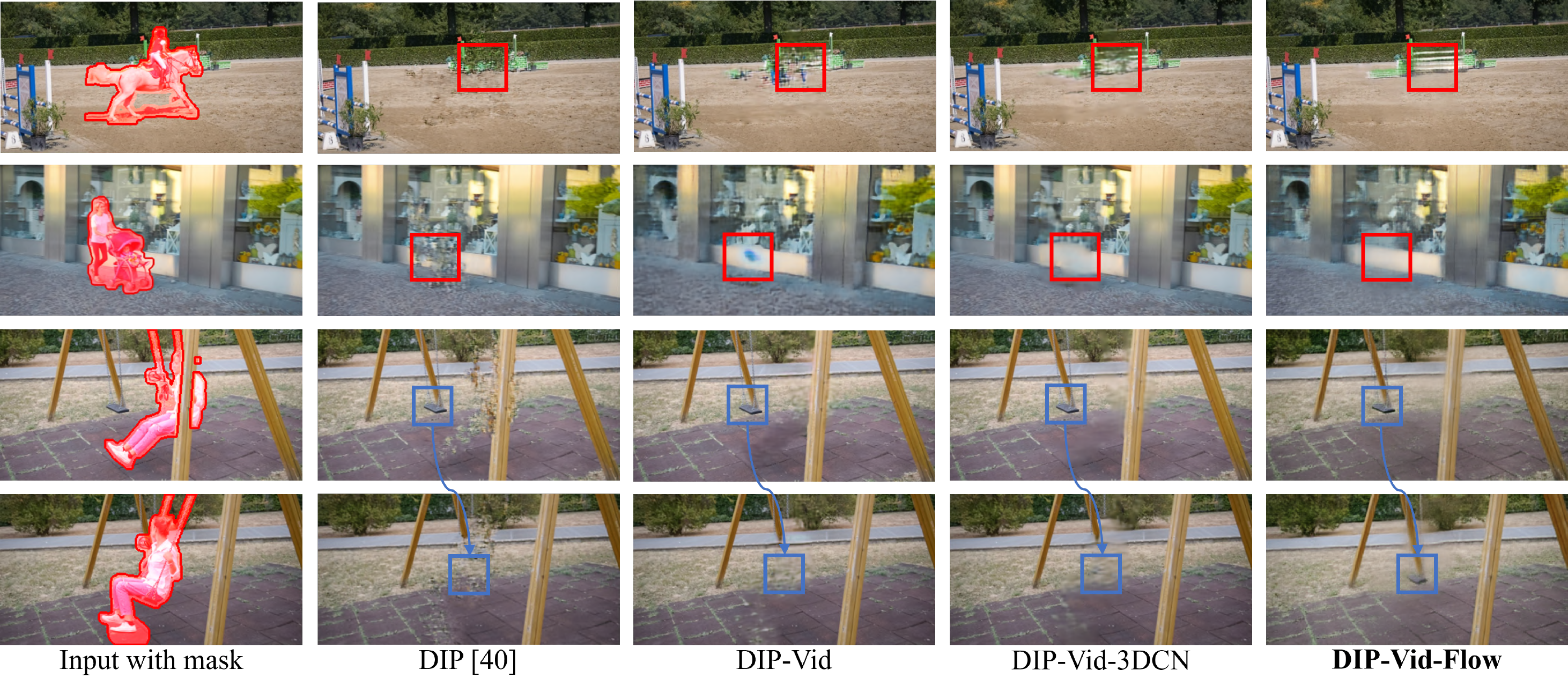}
\end{center}
\vspace{-8mm}
   \caption{Result comparison between different internal learning frameworks (videos provided in~\cite{Huang_SA_2016}). Frame-wise DIP tends to copy textures from known regions, generating incoherent structures. Optimizing over the whole video (DIP-Vid and DIP-Vid-3DCN) improves the visual quality but does not capture temporal consistency well (blue boxes in $3^{rd}$ and $4^{th}$ rows). Our proposed  consistency loss (DIP-Vid-Flow) improves long-term temporal consistency.}
\label{fig:internal_learning}
\vspace{-2mm}
\end{figure*}

We evaluate our method on a variety of real-world videos used in previous works, including 28 videos collected by Huang \etal~\cite{Huang_SA_2016} from the DAVIS dataset~\cite{perazzi2016benchmark}, and 13 videos collected from ~\cite{granados2012background, granados2012not, Newson_SIAM_2014}.

To facilitate quantitative evaluation, we create an additional dataset in which each video has both the foreground masks and the ground-truth background frames. We retrieved 50 background videos from Flickr using different keywords to cover a wide range of scenes and motion types. We randomly select a segment of 60 frames for each video and compose each video with 5 masks randomly picked from DAVIS. This results in 250 videos with real video background and real object masks, which is referred as our Composed dataset.

\subsection{Ablation Study}
\label{sec:ablation_study}

We first compare the video inpainting quality between different internal learning approaches. In particular, we compare our final method, referred to as \textbf{DIP-Vid-Flow}, with the following baselines:

\noindent\textbf{DIP:} This baseline directly applies the `Deep Image Prior' framework~\cite{Ulyanov_CVPR_2018} to video in a frame-by-frame manner.

\noindent\textbf{DIP-Vid:} This is our framework when the model is trained only using the image generation loss (Eq.~\ref{eq:image_loss}). 
% \textcolor{red}{Both DIP and DIP-Vid use the same network architexture as our final method.}

\noindent\textbf{DIP-Vid-3DCN:} Besides directly using the DIP framework in~\cite{Ulyanov_CVPR_2018} with pure 2D convolution, we modify the network to use 3D convolution and apply the image generation loss. 
% Specifically, we treat temporal dimension as the third dimension, and directly convert a 2D batch ($N{\times}{3}{\times}{H}{\times}{W}$) into a 3D batch ($1{\times}3{\times}N{\times}H{\times}W$) with $batchsize=1$.

%We evaluate the video inpainting quality in terms of both frame-wise plausibility and temporal consistency. We measure the visual plausibility of each inpainting frame independently using the Frechet Inception Distance (FID score), a popular metric to measure the quality of images produced by generative models~\cite{heusel2017gans}. We compute the FID score of each inpainted frame independently against the full collection of ground-truth frames from all background videos of our composed dataset. 
\begin{figure*}[t!]\vspace{-0.05in}
\begin{center}
\includegraphics[width=1\linewidth]{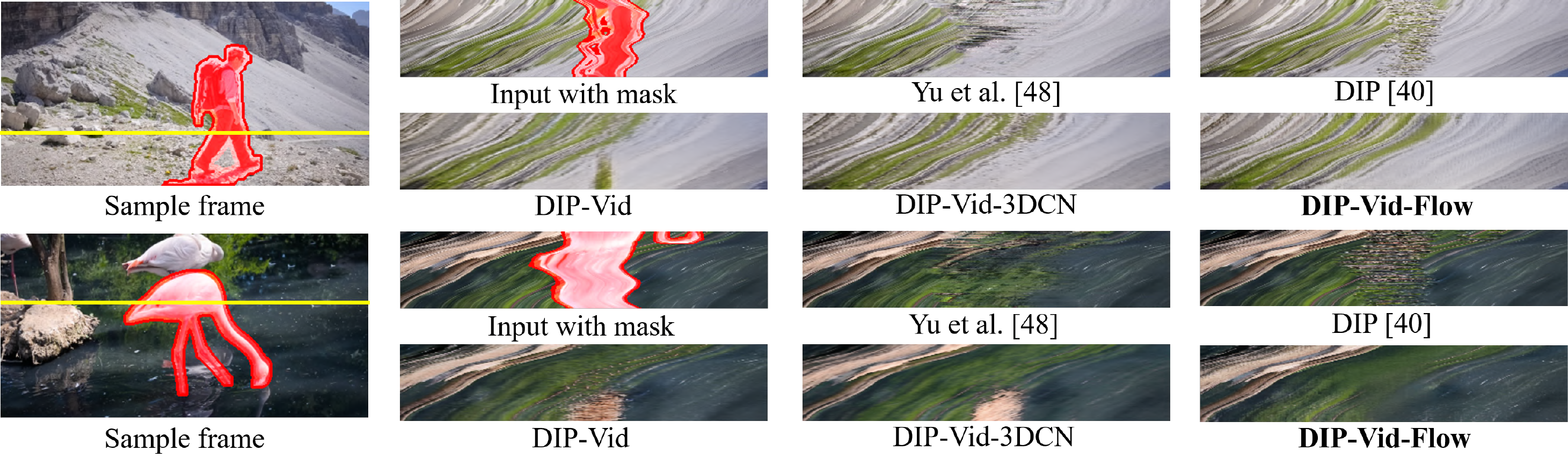}
\end{center}
\vspace{-8mm}
   \caption{Temporal consistency comparison (videos provided in~\cite{Huang_SA_2016}). We stack the pixels in a fixed row (indicated by the yellow line) from all the frames of the video. Our full model (DIP-Vid-Flow) shows the smoothest temporal transition.}
\label{fig:motion_consistency}
\vspace{-4mm}
\end{figure*}

%We evaluate the video inpainting quality in terms of frame-wise visual plausibility and temporal consistency. For visual plausibility, we compute the Fr\'echet Inception Distance (FID) score ~\cite{heusel2017gans} of each inpainted frame independently against the full collection of ground-truth frames from all background videos of our Composed dataset and aggregate the value over the whole video. For motion consistency evaluation, we use the metric introduced in~\cite{gupta2017characterizing}. For each $50 \times 50$ patch sampled in the hole region at frame $t$ we find the patch at time $t + 1$ within 20 pixels of the first patch that maximizes peak signal-to-noise ratio (PSNR) between the two patches. We finally accumulate the PSNR for all the patches and report the average. Similar metric is computed with SSIM~\cite{wang2004image}. 

We evaluate the video inpainting quality in terms of frame-wise visual plausibility, temporal consistency and reconstruction accuracy on our Composed dataset for which the `ground-truth' background videos are available.
For visual plausibility, we compute the Fr\'echet Inception Distance (FID) score~\cite{heusel2017gans} of each inpainted frame independently against the full collection of ground-truth frames and aggregate the value over the whole video.
For temporal consistency, we use the consistency metric introduced in~\cite{gupta2017characterizing}. For each $50{\times}50$ patch sampled in the hole region at frame $t$ we search within the spatial neighborhood of 20 pixels at time $t + 1$ for the patch that maximizes the peak signal-to-noise ratio (PSNR) between the two patches. We compute the average PSNR from all the patches as the metric. Similar metric is also computed using SSIM~\cite{wang2004image}. 
For reconstruction accuracy, we compute the standard PSNR and SSIM on each frame, accumulate the metrics over each video, and report the average performance for each method.

\begin{figure*}[ht]\vspace{-0.1in}
\begin{center}
  \includegraphics[width=1\linewidth]{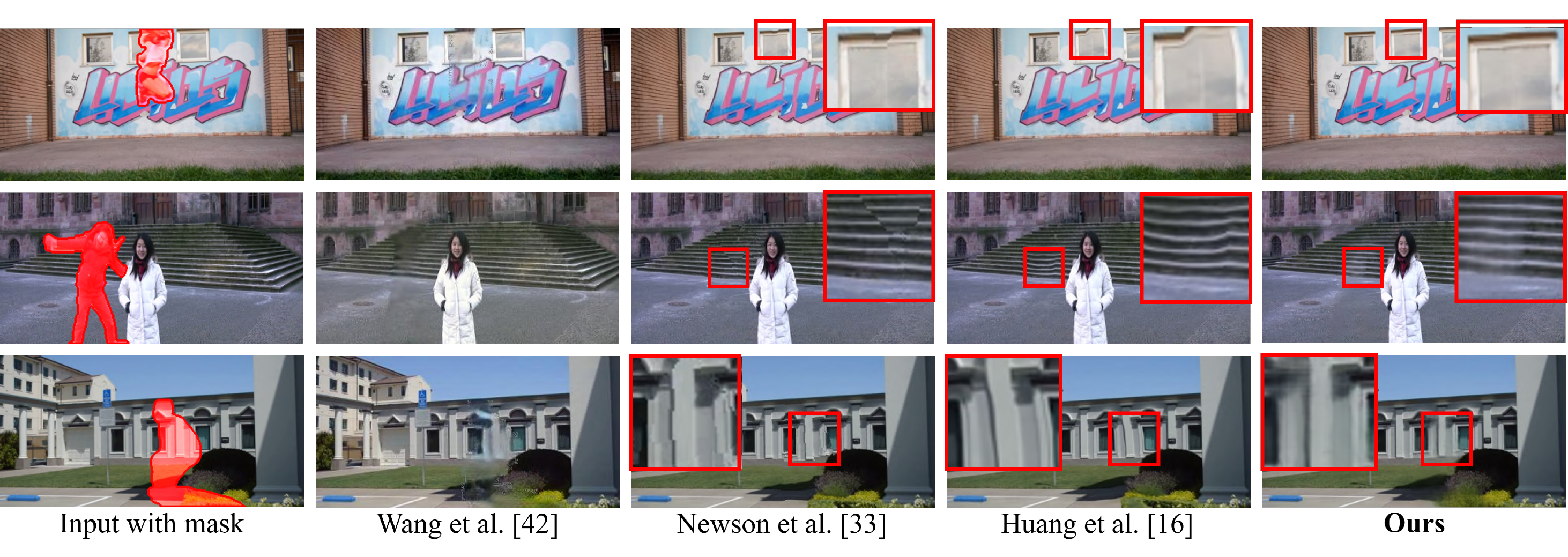}
\end{center}
\vspace{-7mm}
    \caption{Video inpainting results on the videos provided in~\cite{Huang_SA_2016} ($1^{st}$ row) and ~\cite{granados2012background}($2^{nd}$ row) as well as our Composed dataset ($3^{rd}$). Our results are less prone to shape distortion compared to those generated by patch-based methods.}
\label{fig:qualitative}
\vspace{-5mm}
\end{figure*}

% \begin{table}[t!]
% \begin{center}
% \begin{tabular}{|l|c|c|c|c|c|c|}
% \hline
% Method & FID & $\mbox{PSNR}_{\mbox{stab}}$ & $\mbox{SSIM}_{\mbox{stab}}$ \\
% \hline\hline
% % Yu ~\etal~\cite{Yu_CVPR_2018} & 32.20 & 20.5535 & 0.6353 \\
% DIP \cite{Ulyanov_CVPR_2018} & 22.3 & 18.8 & 0.532 \\
% DIP-Vid  & 16.1 & 23.6 & 0.768 \\
% DIP-Vid-3DCN  & 12.1  & 26.7 & 0.871 \\
% DIP-Vid-Flow  & 10.4  & 28.1 & 0.895 \\
% \hline
% \end{tabular}
% \end{center}
% \vspace{-5mm}
%     \caption{Visual plausibility measured in FID and motion consistency measured with PSNR/SSIM between patches of neighboring frames.}
% \vspace{-5mm}
% \label{table:visual_consistency}
% \vspace{-1mm}
% \end{table}

\begin{table}[t!]
\begin{center}
\begin{tabular}{|l|c|c|c|}
\hline
Method & FID & Consistency & PSNR/SSIM\\
% & & (PSNR/SSIM) & \\
\hline\hline
% Yu ~\etal~\cite{Yu_CVPR_2018} & 32.20 & 20.5535 & 0.6353 \\
DIP ~\cite{Ulyanov_CVPR_2018} & 22.3 & 18.8/0.532 & 25.2/.926 \\
DIP-Vid  & 16.1 & 23.6/0.768 & 28.7/.956\\
DIP-Vid-3DCN  & 12.1  & 26.7/0.871 & 30.9/.966\\
\hline
DIP-Vid-Flow  & 10.4  & 28.1/0.895 & 32.1/.969\\
\hline
\end{tabular}
\end{center}
\vspace{-5mm}
    \caption{Ablation Study. Visual plausibility (FID), temporal consistency (PSNR/SSIM), and reconstruction accuracy (PSNR/SSIM) on our Composed dataset. Our full model outperforms all the baselines in all the metrics.}
    % \caption{Visual plausibility score (measured in FID), motion consistency score (measured in PSNR/SSIM) . Our full model (DIP-Vid-Flow) outperforms all the baselines in both metrics.}
\vspace{-5mm}
\label{table:visual_consistency}
\vspace{-1mm}
\end{table}

Tab.~\ref{table:visual_consistency} shows the results of different methods. For all metrics, the video-wise methods significantly improve over the frame-wise DIP method. Incorporating temporal information %into the training framework 
can further improve the results. Explicitly modeling the temporal information in the form of flow prediction leads to the best results.

Fig.~\ref{fig:internal_learning} shows some visual examples. DIP often borrows textures from known regions to fill in the hole, generating incoherent structures in many cases. Training the model over the whole video (DIP-Vid) allows the network to generate better structures, improving the visual quality in each frame. Using 3D convolution tends to constrain the large hole better than 2D due to the larger context provided by the spatial-temporal volume. The result, however, tends to be more blurry and distorted as it is in general very challenging to model the space of spatial-temporal patches. Training the model with our full internal learning framework allows the information to propagate properly across frames which constrains the hole regions with the right information.

%The results are shown in the last two columns of Tab.~\ref{table:visual_consistency}. Applying image inpainting frame-by-frame is highly inconsistent. DIP-Vid, by training the model over whole video, can leverage spatial context similarity across frames to induce motion consistency to some extent but not perfect due to the content mismatch. As expected, DIP-Vid-3DCN improves the motion consistency compared to DIP-Vid. However, as mentioned above, the output space of the 3D CNN is often too large to be learned sufficiently from a single video. We observe that DIP-3DCN tends to mix information from neighboring frames, which results in blurry, misaligned outputs and still often leads to inconsistency (as shown in the third examples ($3^{rd}$ and $4^{th}$ rows) of Fig.~\ref{fig:internal_learning}). With the consistency loss, the correct information is able to be propagated from neighboring frames allow for better consistency with a 2D CNN.

Fig.~\ref{fig:motion_consistency} visualizes temporal consistency of different video inpainting results on two video sequences.% ``hike'' and ``flamingo'' in DAVIS. 
We visualize the video content at a fixed horizontal stride across the whole video. Note that the strides cut through the hole regions in many frames. As the video progresses, the visualization from a good video inpainting result should appear smooth. Applying the image inpainting methods~\cite{Ulyanov_CVPR_2018, Yu_CVPR_2018} result in inconsistency between the hole regions and non-hole regions across the video. DIP-Vid and DIP-Vid-3DCN result in smoother visualizations compared to DIP yet still exhibit inconsistent regions while our full model gives the smoothest visualization, indicating high temporal consistency.  

\subsection{Video Inpainting Performance}
\label{sec:video_performance}

% We evaluate the inpainting performance of our internal-learning framework and compare with different state-of-the-art inpainting methods. For quantitative evaluation, we evaluate the inpainting results on our Composed dataset for which the ``ground-truth'' background videos are available. In addition to the baseline methods described in Sec.~\ref{sec:ablation_study}, we also include the inpainting results obtained from one state-of-the art image inpainting method (Yu \etal~\cite{Yu_CVPR_2018}), the Vid2Vid (Wang \etal~\cite{wang2018vid2vid}) model trained on video inpainting data, and two state-of-the-art video inpainting methods by Newson \etal~\cite{Newson_SIAM_2014} and Huang \etal~\cite{Huang_SA_2016}. 
We compare the video inpainting performance of our internal-learning method with different state-of-the-art inpainting methods, including the inpainting results obtained from one state-of-the art image inpainting method by Yu \etal~\cite{Yu_CVPR_2018}, Vid2Vid (Wang \etal~\cite{wang2018vid2vid}) model trained on video inpainting data, and two state-of-the-art video inpainting methods by Newson \etal~\cite{Newson_SIAM_2014} and Huang \etal~\cite{Huang_SA_2016}.
For Vid2Vid, we train the model on a different composed dataset created for video inpainting. The training set containing 1000 videos of 30-frames each is constructed with the same procedure used for our Composed dataset.

\begin{table}
\begin{center}
\begin{tabular}{|l|c|c|c|}
\hline
\diagbox[width=9em]{Method}{PSNR/SSIM} & All & Complex & Simple \\
\hline\hline
Yu \etal~\cite{Yu_CVPR_2018} & 24.9/.929 & 24.7/.926 & 25.1/.931\\
Wang \etal~\cite{wang2018vid2vid} & 26.0/.914 & 25.3/.908 & 26.6/.920\\
Newson \etal~\cite{Newson_SIAM_2014} & 30.6/.962 & 30.9/.963 & 30.4/.960\\
Huang \etal~\cite{Huang_SA_2016} & 32.3/.971 & 31.6/.968 & 33.0/.974\\
% \hline
% DIP ~\cite{Ulyanov_CVPR_2018} & 25.2/.926 & 25.0/.925 & 25.4/.927\\
% DIP-Vid  & 28.7/.956 & 28.7/.957 & 28.7/.955\\
% DIP-Vid-3DCN  & 30.9/.966 & 31.2/.967 & 30.7/.965\\
\hline
Ours & 32.1/.969 & 31.9/.970 & 32.2/.968\\
\hline
\end{tabular}
\end{center}
\vspace{-5mm}
    \caption{Quantitative Evaluation. PSNR/SSIM on our Composed dataset as well as its two partitions. Our method produces more accurate video inpainting results than most of the existing methods. Our method performs favorably for challenging videos with complex motion.}
\vspace{-5mm}
\label{table:quantitative}
\end{table}

% Using the ground-truth background videos, we evaluate the inpainting results with the standard PSNR and SSIM for each frame, accumulate the metrics over each video, and report the average performance for each method in Tab.~\ref{table:quantitative}. 
For quantitative evalution, we report PSNR and SSIM on our Composed dataset in Tab.~\ref{table:quantitative}. The results show that our method produces more accurate video inpainting results than most of the existing %video inpainting 
methods, except for Huang \etal~\cite{Huang_SA_2016}. Fig.~\ref{fig:qualitative} shows some visual examples of the inpainted frames from different methods. Please refer to our supplementary video for more results.

Interestingly, we observe that our results complement those of the patch-based methods when videos have more complex background motion or color/lighting changes. Patch-based methods rely on explicit patch matching and flow tracking during synthesis, which often fail when the appearance or motion varies significantly. This leads to distorted shapes inconsistent with the known regions (see Fig.~\ref{fig:qualitative} for visual examples). On the other hand, our network-based synthesis tends to capture better natural image priors and handles those scenarios more robustly.

\begin{figure}[t!]
% \centering
% \subfloat{
%   \includegraphics[width=0.58\linewidth]{latex/figures/fig7/comparison_PSNR.pdf}
% }\hspace*{-1.5em}
% \subfloat{
%   \includegraphics[width=0.58\linewidth]{latex/figures/fig7/comparison_SSIM.pdf}
% }
\begin{center}
\includegraphics[width=1.0\linewidth]{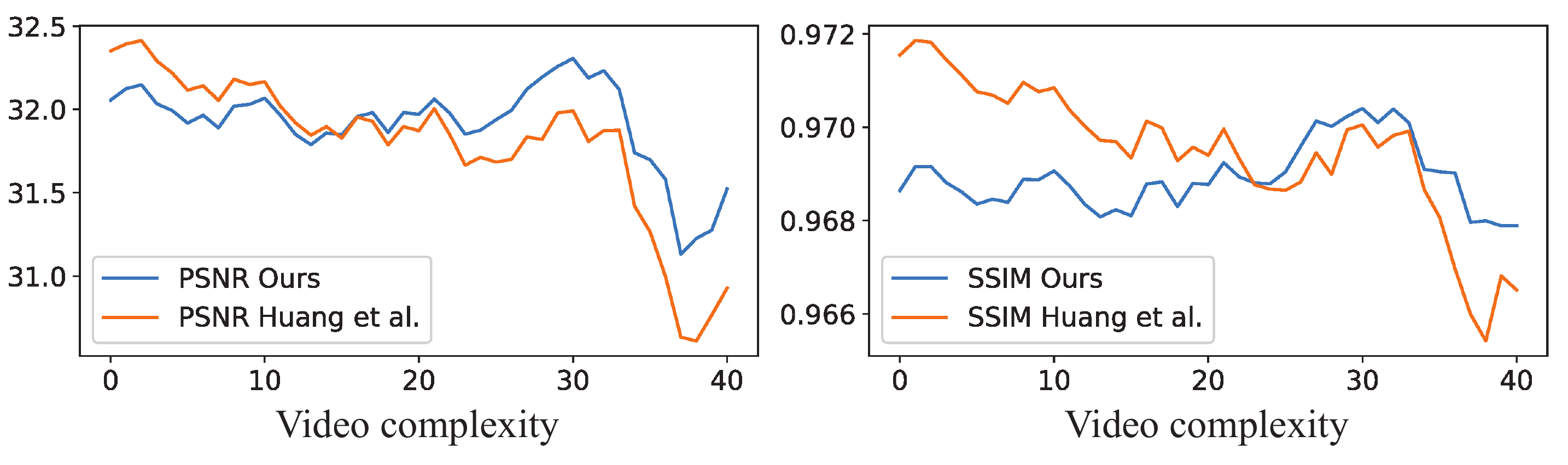}
\end{center}
\vspace{-6mm}
    \caption{Comparison between our method and the state-of-the-art video inpainting method~\cite{Huang_SA_2016}. We plot the average performance of videos whose complexity metric are above the threshold in x-axis. Our method performs consistently better when the appearance and motion gets more complex.}
\label{fig:comparison}
\vspace{-3mm}
\end{figure}

To further understand this complementary behavior, we separate our Composed dataset into two equal partitions according to video complexity defined by the metric as the sum of two terms: the standard deviation of average pixel values across the frames, which measures the appearance change; and the mean of the flow gradient magnitude, which reflects the motion complexity.
We report PSNR and SSIM for these two partitions in Tab.~\ref{table:quantitative}, as well as in Fig.~\ref{fig:comparison}, where we plot the average performance of videos whose complexity metric is above certain threshold. It shows that our method tends to perform slightly better compared with patch-based methods when the video complexity increases.

\begin{figure}[t!]
\begin{center}
% \fbox{\rule{0pt}{2in} \rule{.9\linewidth}{0pt}}
\includegraphics[width=1\linewidth]{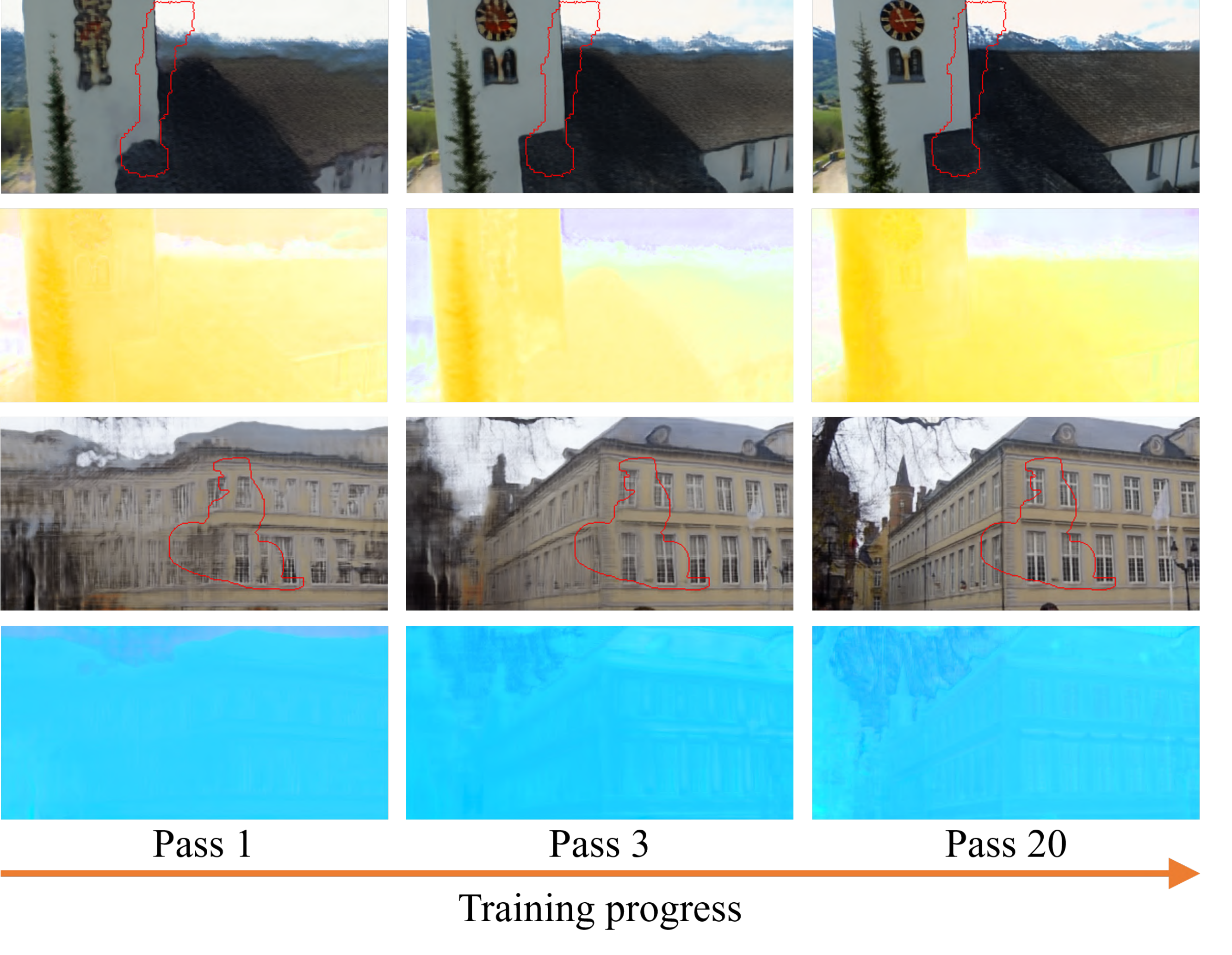}
\vspace{-8mm}
   \caption{Intermediate results during training. Generated frames: red curve outlines the hole region. Generated flow maps (forward flow with range 1): hue denotes the flow orientation, value denotes the magnitude. As the training progresses, the model captures more accurate texture and motion (the flow map  becomes smoother and more consistent with the actual video scene).}
\vspace{-3mm}
\label{fig:training_time}
\vspace{-2mm}
\end{center}
\end{figure}

Fig.~\ref{fig:training_time} shows our generated frames and flow maps during training process. After one pass through all the frames, our model is able to reconstruct the global structure of the frames. Ghosting effect is usually observed at this stage because the mapping from noise to image is still not well established for individual frames. During the next several passes, the model gradually learns more texture details of the frames, both inside and outside the hole region. 
The inpainting result is generally good after just a few passes through the whole video. In practice, we keep training longer to further improve the long-term temporal consistency, since it takes more iterations for the contents from distant frames to come into play via the consistency loss. 
In Fig.~\ref{fig:training_time}, we observe that the predicted flow in the hole region is coherent with their neighboring content, indicating effective flow inpainting. We also note that the predicted flow orientation (represented by the hue in the visualized flow maps) is also consistent with the camera motion, thereby serving as a guidance for the image generation.
% \textcolor{red}{Although our final generated flow maps sometimes are not as smooth as the estimated flows, the flow in the hole region maintains coherent with its neighboring content and also consistent with the camera motion, which is sufficient for guiding the image generation.}

%% file: latex/Discussion.tex
% In this section, we visualize the encoded latent feature to understand how our learned encoding 
% tracks the same image content through the video, and investigate the influence of input window length to further understand our model.
In this section, we visualize the encoded latent feature and investigate the influence of input window length to further understand our model.

\subsection{Visualization of Encoded Latent Feature}

\begin{figure*}[h!]
\begin{center}
\includegraphics[width=0.99\linewidth]{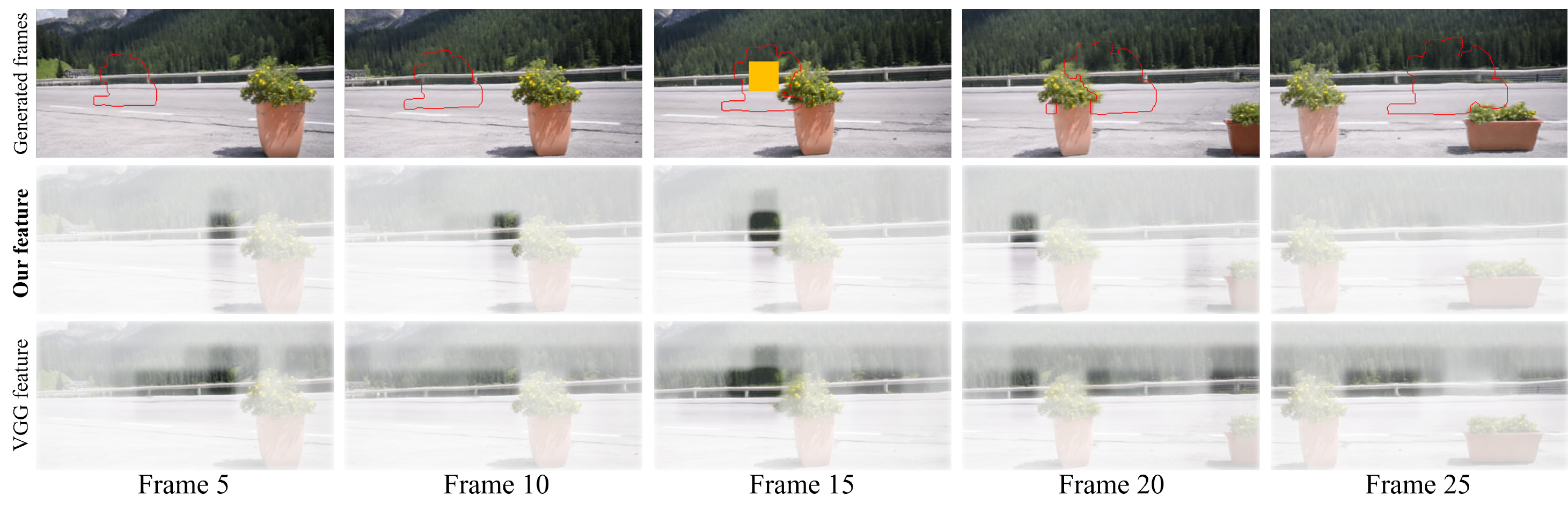}
\end{center}
\vspace{-8mm}
   %\caption{Visualization of encoded feature similarity: We pick a patch from the frame shown in the middle, then compute similarity score between this patch and all the other patches across the five frames, and visualize the similarity score with the alpha channel. The two features we used: (middle)feature encoded by our model; (bottom)feature encoded by VGG16-pool5. Our encoded feature has the potential to track identity instead of pure visual similarity.}
   %\caption{Feature similarity visualization. The alpha values represent the similarity between each patch to the referenced patch (yellow square) in the encoder feature space.
   %Top: original frames with outlined hole regions.
   %Middle: visualization using our feature (encoded from the input noise).
   %Bottom: Visualization using VGG16-pool5 feature (encoded from our generated frames).
   %Our learned feature tends to identify the exact patch instead of just searching for visually similar ones.}
   \caption{Visualization of feature similarity of our model compared with VGG. A higher opacity patch indicates higher feature similarity. Our learned feature is able to track the exact patch instead of only searching for visually similar patches.}
\label{fig:longterm_consistency}
\vspace{-5mm}
\end{figure*}

Recall that our generator $G_{\theta}$ has an Encoder-Decoder structure that encodes random noise to latent features and decodes the features to generate image pixels. To better understand how the model works, it is helpful to inspect the feature learned by the encoder. In Fig.~\ref{fig:longterm_consistency}, we visualize the feature similarity between a reference patch and all the other patches from neighboring locations and other frames. Specifically, we select a reference patch from the middle frame with patch size 32x32 matching the receptive field of the encoder, and calculate the cosine similarity between the features of the reference and other patches. The feature vector of a patch is given by the neuron responses at the corresponding 1x1 spatial location of the feature map. The patch similarity map is shown on the middle frame as well as 4 nearby frames, encoded in the alpha channel. A higher opacity patch indicates a higher feature similarity. As a comparison, we show the similarity map calculated with both our learned feature (middle row) and VGG16-pool5 feature (bottom row). It can be observed from the example in Fig.~\ref{fig:longterm_consistency} that the most similar patches identified by our learned feature are located on the exact same object across a long range of frames; VGG feature can capture general visual similarity but fails to identify the same object instance. 
This interesting observation provides some indication that certain video specific features have been learned during the internal learning process.
\subsection{Influence of Window Length}
\label{sec:insights}

\begin{figure}[t!]
% \centering
% \subfloat{
%   \includegraphics[width=0.58\linewidth]{latex/figures/fig8/clip_length_psnr.pdf}
% }\hspace*{-1.5em}
% \subfloat{
%   \includegraphics[width=0.58\linewidth]{latex/figures/fig8/clip_length_ssim.pdf}
% }
\begin{center}
\vspace{-2mm}
\includegraphics[width=1.0\linewidth]{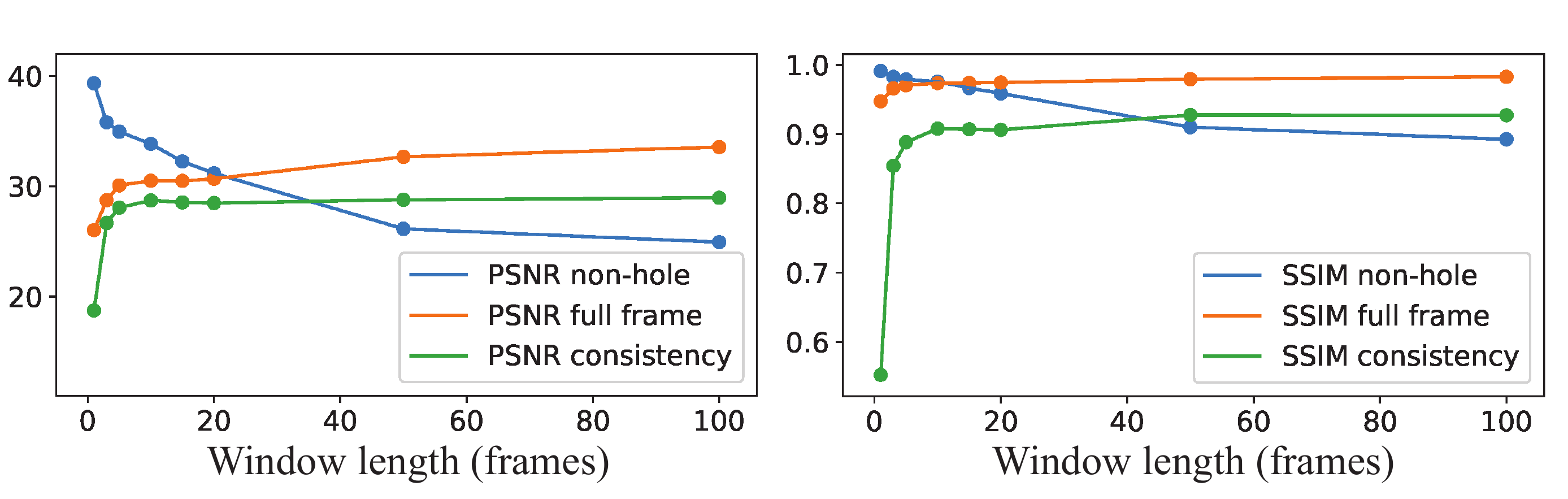}
\end{center}
\vspace{-6mm}
    \caption{Influence of window length: As the window length increases, the reconstruction quality in the non-hole region decreases. However, the overall inpainting quality improves, indicating the improvement in generalization in the hole regions as more frames become available.}
\vspace{-5mm}
\label{fig:window_length}
\end{figure}

%Due to the limited capacity of the model, it is impractical to train the model over a very long video sequence. At the same time, the information to fill the hole mostly comes from the local neighboring frames. Therefore, for long video sequence, it is reasonable to cut it into multiple small clips and train a model to fit each of the clip, which can also be easily parallelized. In this section, we experiment with different clip length on a subset of composed dataset. 

In our framework, the input video can be considered as the training data with which the inpainting generative network is trained. To investigate how the inpainting performance is affected by the input window length, we perform an experiment on a subset of 10 videos randomly selected from our Composed dataset. We divide each video into clips of window length $k$ and apply our inpainting method on each clip independently. We experiment with different window length $k \in \{1, 3, 5, 10, 15, 25, 50, 100\}$. We plot the average PSNR and SSIM scores of both generated non-hole region and full frame (the generated hole fused with the input non-hole region), as well as the consistency metric introduced in Sec. \ref{sec:ablation_study} under each setting in Fig.~\ref{fig:window_length}.

Interestingly, while the overall inpainting quality improves as $k$ increases, the reconstruction quality in the non-hole region decreases. This indicates overfitting when training on limited data. In fact, when $k{=}1$, it reduces to running DIP frame-wise which gives the best reconstruction quality on the non-hole region. The fact that our method achieves better inpainting results with larger window length indicates the capability of the model in leveraging distant frames to generate better results in the hole region.

%In practice, we value more of the quality inside the hole since we can always fuse the inpainted hole region back to the input non-hole region.

%When $k = 1$, it reduces to running DIP frame-wise, which gives the highest reconstruction quality on non-hole region. Meanwhile, with the increase in clip length, our model generates more visually plausible and temporal consistent results inside the hole, which leads to improvement in the full-frame measure. In practice, we value more of the quality inside the hole since we can always fuse the inpainted hole region back to the input non-hole region. However, with larger clip length, the drop of the reconstruction quality also indicates that it is harder for the model to generalize well when the content of the video changes a lot, and this also potentially affects the inpainting quality inside the hole (e.g. more blurry).    

\subsection{Limitation and Future Work}

As is the case with DIP~\cite{Ulyanov_CVPR_2018} and many other back-propagation based visual synthesis system, long processing time is the main limitation of our method. It often takes hours to train an individual model for each input video. 
Our method can fail when the hole is large and has little motion relative to the background. In those cases, there is too little motion to propagate the content across frames.
%The main limitation of our framework is the long processing time. For each input video, our framework need to train a separate model, which often take hours to finish. That makes it not suitable for practical video inpainting systems.
Nevertheless, the value of our work lies in exploring the possibilities of internal learning on video inpainting and identifying its strength that complements other learning-based methods relying on external training data.
In future work, we plan to further investigate how to combine representations learned internally with externally trained models to enable powerful learning systems.
% To make our approach more practical, we can further investigate how to combine representations learned internally with externally trained models to enable powerful learning systems. 
%we can further investigate more efficient optimization algorithms and distributed computing solutions. 

%main goal in this paper is not to develop another state-of-the-art video inpainting method. We instead focus on 
%We hope that the findings and insights in this work can encourage future works to explore this interesting research direction. For example, combining  representations learned internally with externally trained models can potentially enable  powerful learning systems.

It remains an open question that, in the context of internal learning, what network structure can best serve as a prior to represent video sequence data. In this work, we have intentionally restricted the network to a 2D CNN structure to study the capability of such a simple model in encoding temporal information. In future work, we plan to study more advanced architectures with explicit in-network temporal modeling, such as recurrent networks and sequence modeling in Vid2Vid~\cite{wang2018vid2vid}. 
%Video generation necessarily requires the temporal structure of video data to be modeled. In this work, we intentionally keep the simplicity of the model at an extreme end (\ie frame-wise prediction with a 2D CNN) to stress its ability in capturing the temporal structure via internal learning. To better handle temporal consistency, however, we believe that more advanced architectures with explicit in-network temporal modeling can be adapted. In future works, we plan to  explore adapting more advanced video generation architecture such as Vid2Vid in the context of internal learning.

%% file: latex/Conclusion.tex
In this paper, we introduce a novel approach for video inpainting based on internal learning. In extending Deep Image Prior ~\cite{Ulyanov_CVPR_2018} to video, we explore effective strategies for internal learning to address the fundamental challenge of temporal consistency in video inpainting. We propose a consistency-aware training framework to jointly generate both appearance and flow, whilst exploiting these complementary modalities to ensure mutual consistency. 
We demonstrate that it is possible for a regular image-based generative CNN to achieve coherent video inpainting results, while optimizing directly on the input video without reliance upon an external corpus of visual data. With this work, we hope to attract more research attention to the interesting direction of internal learning, which is  complementary to the mainstream large-scale learning approaches. We believe combining the strengths from both directions can potentially lead to better learning methodologies.

\noindent \textbf{Acknowledgments} This project is supported in part by the Brown Institute for Media Innovation. We thank Flickr users Horia Varlan, tyalis\_2, Andy Tran and Ralf Kayser for their permissions to use their videos in our experiments.

%% file: latex/Appendix.tex
\section{Network Architecture}
In our experiments, we use two different networks. Our 2D baselines (DIP and DIP-Vid) and our final model (DIP-Vid-Flow) share the same Encoder-Decoder architecture. Our 3D baseline (DIP-Vid-3DCN) uses a modified version with 3D convolution. The source code is available at our project website \url{https://cs.stanford.edu/~haotianz/publications/video_inpainting/}.

\subsection{2D Network}
\noindent \textbf{Encoder}
The Encoder consists of 12 convolution layers. Every two consecutive layers form a block, where the two layers have the same number of channels. The first layer in each block uses the stride of 2 to reduce the spatial resolution. All the convolution layers use the filter size of 5. The number of channels for each layer is shown below.

\noindent C16-C16-C32-C32-C64-C64-C128-C128-C128-C128-C128-C128

\noindent \textbf{Decoder}
The Decoder also consists of 12 convolution layers in 6 blocks. One Nearest-neighbor upsampling layer is added to the beginning of each block. All the convolution layers use the filter size of 3. The number of channels for each layer are symmetric to those in the Encoder.

\noindent \textbf{Skip Connection}
A skip connection is added from the beginning of the $i$ th block of the Encoder to the beginning of the $(n-i)$ th block (after the upsampling layer) of the Decoder. All skip connections use one convolution layer with 4 channels and filter size of 1.

\noindent \textbf{Final Layer}
For DIP and DIP-Vid, the final layer only contains a convolution layer with 3 channels followed by a sigmoid to generate the final image. For DIP-Vid-Flow, a flow generation branch is added parallel to the image generation branch, which contains a convolution layer with 12 channels, corresponding to 6 different flow maps of temporal range 1, 3, 5 in both forward and backward directions. 

All the convolution layers except those in the final layer are followed by a Batch-Norm layer and a LeakyReLU layer with slope 0.2.

\subsection{3D Network}
Our 3D version of the network shares the same structure with the 2D version except all the 2D convolution layers are replaced with 3D convolution layers. We also keep all the number of channels and filter size as the same as our 2D version. For the added $3^{rd}$ dimension, we use the filter size of 3 for the Encoder and the Decoder and the filter size of 1 for the skip connections.

\section{Network Input}
As mentioned in the main paper, we sample the input noise maps independently for each frame and fix them during training. The noise map has one channel and shares the same spatial size with the input frame. Each noise map is filled with uniform noise between 0 and 0.1. For our 2D network, we feed input noise maps as a 2D batch of dimension $N \times 1 \times H \times W$, where N is the batch size. For our 3D network, we transfer the 2D batch into a 3D batch of dimension $1 \times 1 \times N \times H \times W$, where $N$ becomes the size of the $3^{rd}$ dimension with batch size of 1. In all of our experiments, frames are resized and cropped to $384 \times 192$.

\section{Network Training}
In this section, we describe the training details for all the baselines and our final model.

\subsection{DIP}
We train a DIP model for each frame independently. Due to the destabilization issue mentioned in the original paper ~\cite{Ulyanov_CVPR_2018}, we run optimization on each frame for 5k iterations and save the result every 100 iterations. The intermediate result with the lowest loss is chosen as the final result. 

\subsection{DIP-Vid}
We train a single DIP model on the entire video. In each epoch, we randomly pick $N$ consecutive frames as a training batch to enumerate all the possible batch permutations. Inspired by the training procedure used in DIP, we run optimization on the selected batch for $M$ iterations before moving to the next batch. After training for $E$ epochs, we run one inference using the trained model to get the final inpainting results. Only the image generation loss is applied in this baseline. The destabilization issue is also observed in this method, but considerably rare compared to DIP.

\subsection{DIP-Vid-3DCN}
All the settings are as the same as DIP-Vid, except for replacing the 2D network with the 3D version.

\subsection{DIP-Vid-Flow}
In our final model, we need to generate both images and flows. We randomly pick N frames which are
consecutive with a fixed frame interval of t as a batch, $t \in \{1, 3, 5\}$. We do not use intervals larger than 5 due to the increasing error in estimated flows. We run optimization on the batch with all the image and flow related loss (See Sec3.1 in our main paper), but only using forward or backward flow at interval $t$ for $M$ iterations. Optimizing flows in both directions at the same time is observed to cause artifacts in the results occasionally, potentially due to the conflict in the flows.   
We select batches by enumerating all the possible permutations and finish training after $E$ epochs on the whole video. 

We use $N=5$, $M=100$ and $E=20$ in all of our experiments.